\documentclass[11pt,a4paper]{article}
\usepackage[hyperref]{emnlp-ijcnlp-2019}
\usepackage{times}
\usepackage{amsmath}
\usepackage{latexsym}
\usepackage{algorithm,algorithmic}
\usepackage{url}

\usepackage{amsmath,amsfonts,amssymb,bm}
\usepackage{pifont} 
\usepackage{graphicx,subfigure,epsfig,fancybox} 
\usepackage{float}
\usepackage{color} 
\usepackage{multirow}
\usepackage{natbib}

\renewcommand{\vec}[1]{\boldsymbol{#1}}

\newcommand{\mat}[1]{\mathbf{#1}}

\DeclareMathOperator*{\argmax}{argmax}

\usepackage{adjustbox}
\usepackage{array}
\usepackage{booktabs}

\newcolumntype{R}[2]{%
    >{\adjustbox{angle=#1,lap=\width-(#2)}\bgroup}%
    l%
    <{\egroup}%
}

\newcommand{\softmax}[1]{\text{softmax}\left(#1\right)}

\usepackage{tikz}
\usetikzlibrary{shapes.geometric, positioning, calc}
\usetikzlibrary{arrows,shapes,calc}
\usetikzlibrary{trees,positioning,mindmap,shadows,fit}
\usetikzlibrary{decorations.pathreplacing}
\usetikzlibrary{tikzmark} 
\usetikzlibrary{intersections} 

\aclfinalcopy 

\title{An Empirical Comparison on Imitation Learning and Reinforcement Learning for Paraphrase Generation}

\author{Wanyu Du \\
  Department of Computer Science \\
  University of Virginia \\
  Charlottesville, VA 22904 \\
  {\tt wd5jq@virginia.edu} \\\And
  Yangfeng Ji \\
  Department of Computer Science \\
  University of Virginia \\
  Charlottesville, VA 22904 \\
  {\tt yangfeng@virginia.edu} \\}

\date{}

\begin{document}
\maketitle
\begin{abstract}
Generating paraphrases from given sentences involves decoding words step by step from a large vocabulary.
To learn a decoder, supervised learning which maximizes the likelihood of tokens always suffers from the exposure bias.
Although both reinforcement learning (RL) and imitation learning (IL) have been widely used to alleviate the bias, the lack of direct comparison leads to only a partial image on their benefits.
In this work, we present an empirical study on how RL and IL can help boost the performance of generating paraphrases, with the pointer-generator as a base model \footnote{The data and code for this work can be obtained from: \url{https://github.com/ddddwy/Reinforce-Paraphrase-Generation}}.
Experiments on the benchmark datasets show that (1) imitation learning is constantly better than reinforcement learning; and (2) the pointer-generator models with imitation learning outperform the state-of-the-art methods with a large margin.
\end{abstract}

\section{Introduction}
\label{sec:intro}

Generating paraphrases is a fundamental research problem that could benefit many other NLP tasks, such as machine translation~\citep{bahdanau2014neural}, text generation~\citep{radford2019language}, document summarization~\citep{chopra2016abstractive}, and question answering~\citep{mccann2018natural}.
Although various methods have been developed \citep{Zhao:2009:ASP:1690219.1690263, quirk2004monolingual, Barzilay:2003:LPU:1073445.1073448}, the recent progress on paraphrase generation is mainly from neural network modeling~\citep{prakash2016neural}.
Particularly, the encoder-decoder framework is widely adopted~\citep{cho2015describing}, where the encoder takes source sentences as inputs and the decoder generates the corresponding paraphrase for each input sentence.

In supervised learning, a well-known challenge of generating paraphrases is the exposure bias: the current prediction is conditioned on the ground truth during training but on previous predictions during decoding, which may accumulate and propagate the error when generating the text. 
To address this challenge, prior work \citep{li-etal-2018-paraphrase} suggests to utilize the exploration strategy in reinforcement learning (RL).
However, training with the RL algorithms is not trivial and often hardly works in practice~\citep{dayan2008reinforcement}.
A typical way of using RL in practice is to train the model with supervised learning~\citep{ranzato2015sequence, shen2015minimum, bahdanau2016actor}, which leverages the supervision information from training data and alleviate the exposure bias to some extent.
In the middle ground between RL and supervised learning, a well-known category is imitation learning (IL)~\citep{daume2009search,ross2011reduction}, which has been used in structured prediction~\citep{bagnell2007boosting} and other sequential prediction tasks~\citep{bengio2015scheduled}.\footnote{In this work, we view \emph{scheduled sampling} \citep{bengio2015scheduled} as an imitation learning algorithm similar to \textsc{Dagger}~\citep{ross2011reduction}.}

In this work, we conduct an empirical comparison between RL and IL to demonstrate the pros and cons of using them for paraphrase generation.
We first propose a unified framework to include some popular learning algorithms as special cases, such as the \textsc{Reinforce} algorithm~\citep{williams1992simple} in RL and the \textsc{Dagger} algorithm~\citep{ross2011reduction} in IL.
To better understand the value of different learning techniques, we further offer several variant learning algorithms based on the RL framework.
Experiments on the benchmark datasets show: (1) the \textsc{Dagger} algorithm is better than the \textsc{Reinforce} algorithm and its variants on paraphrase generation, (2) the \textsc{Dagger} algorithm with a certain setting gives the best results, which outperform the previous state-of-the-art with about 13\% on the average evaluation score.
We expect this work will shed light on how to choose between RL and IL, and alleviate the exposure bias for other text generation tasks.

\section{Method}
\label{sec:methods}

Given an input sentence $\vec{x} = (x_1, x_2, \cdots, x_S)$ with length $S$, a paraphrase generation model outputs a new sentence $\vec{y} = (y_1, y_2, \cdots, y_T)$ with length $T$ that shares the same meaning with $\vec{x}$.
The widely adopted framework on paraphrase generation is the encoder-decoder framework \citep{cho2014learning}.
The encoder reads sentence $\vec{x}$ and represents it as a single numeric vector or a set of numeric vectors. 
The decoder defines a probability function $p(y_{t}\mid \vec{y}_{\leq t-1},\vec{x};\vec{\theta})$, where $\vec{y}_{\leq t-1}=(y_1,y_2,\dots,y_{t-1})$ and $\vec{\theta}$ is the collection of model parameters,
\begin{equation}
  \label{eq:rnn}
  p(y_{t}\mid \vec{y}_{\leq t-1},\vec{x};\vec{\theta})=\softmax{\mat{W}\vec{h}_{t}}
\end{equation}
with $\vec{h}_{t}=\vec{f}(\vec{h}_{t-1},y_{t-1},\vec{x})$, where $\vec{f}$ as a nonlinear transition function and $\mat{W}\in\vec{\theta}$ as a parameter matrix.
We use the \emph{pointer-generator model} \citep{see2017get} as the base model, which is state-of-the-art model on paragraph generation~\citep{li-etal-2018-paraphrase}.
We skip the detail explanation of this model and please refer to \citep{see2017get} for further information.

\subsection{Basic Learning Algorithms}
\label{subsec:learning}

To facilitate the comparison between RL and IL, we propose a unified framework with the following objective function.
Given a training example $(\vec{x},\vec{y})$, the objective function is defined as
\begin{equation}
  \label{eq:obj}
  L(\vec{\theta}) = \big\{
  \sum_{t=1}^{T}\log \pi_{\theta}(\tilde{y}_t\mid \vec{h}_{t})
  \big\}\cdot r(\tilde{\vec{y}},\vec{y}),
\end{equation}
Following the terminology in RL and IL, we rename $P(\tilde{y}_t\mid \vec{y}_{\leq t-1},\vec{x};\vec{\theta})$ as the the policy function $\pi_{\theta}(\tilde{y}_t\mid \vec{h}_t)$.
That implies taking an action based on the current observation, where the action is \emph{picking} a word $\tilde{y}_t$ from the vocabulary $\mathcal{V}$. 
$r(\tilde{\vec{y}},\vec{y})$ is a reward function with $r(\tilde{\vec{y}},\vec{y})=1$ if $\tilde{\vec{y}}=\vec{y}$.
In our experiments, We use the ROUGE-2 score \citep{lin2004rouge} as the reward function.
Algorithm \autoref{alg:learning} presents how to optimize $L(\vec{\theta})$ in the online learning fashion.
As shown in the pseudocode, the {\bf schedule rates $(\alpha,\beta)$} and the {\bf decoding function} $\texttt{Decode}(\cdot)$ are the keys to understand the special cases of this unified framework.

\begin{algorithm}
  \caption{Online learning}
  \label{alg:learning}
  \begin{algorithmic}[1]
    \STATE {\bf Input}: A training example $(\vec{x}^{(i)},\vec{y}^{(i)})$, current schedule rates $\alpha^{(i)},\beta^{(i)} \in[0,1]$, learning rate $\eta$
    \STATE Initialize $L(\vec{\theta})\leftarrow 0$
    \FOR {$t = 1,\dots,T$}
    \STATE $p_1,p_2\sim\text{Uniform}(0,1)$
    \STATE $\tilde{y}_{t-1}\leftarrow y_{t-1}~\textbf{if}~(p_1 < \alpha^{(i)})~\textbf{else}~\hat{y}_{t-1}$
    \STATE $\vec{h}_{t}=\vec{f}(\vec{h}_{t-1},\tilde{y}_{t-1},\vec{x})$
    \STATE $\hat{y}_{t}\leftarrow \texttt{Decode}(\pi(y\mid \vec{h}_{t}))$
    \STATE $\tilde{y}_{t}\leftarrow y_{t}~\textbf{if}~(p_2 < \beta^{(i)})~\textbf{else}~\hat{y}_{t}$
    \STATE $L(\vec{\theta})\leftarrow L(\vec{\theta}) + \log\pi(\tilde{y}_{t}\mid \vec{h}_t)$
    \ENDFOR
    \STATE $\delta\vec{\theta}\leftarrow\nabla_{\vec{\theta}}L(\vec{\theta})\cdot r(\tilde{\vec{y}^{(i)}},\vec{y}^{(i)})$
    \STATE $\vec{\theta}\leftarrow \vec{\theta} + \eta\cdot\delta\vec{\theta}$
  \end{algorithmic}
\end{algorithm}

\paragraph{The \textsc{Reinforce} Algorithm.}
When $\alpha=0$, $\beta=0$, and $\texttt{Decode}(\pi(y\mid\vec{h}_t))$ is defined as as:
\begin{equation}
  \label{eq:sampling}
  \begin{split}
    &\texttt{Decode}(\pi_{\theta}(y\mid \vec{h}_{t-1}))\\
    & ~~~=\text{Random\_Sampling}(\pi_{\theta}(y\mid \vec{h}_{t-1})),
  \end{split}
\end{equation}
Specifically, when $\alpha=\beta=0$, both $\tilde{y}_{t-1}$ and $\tilde{y}_{t}$ will choose the sampled values from the \texttt{Decode} function with policy $\pi_{\theta}$.
It essentially samples a trajectory from the decoder $(\hat{y}_1,\hat{y}_2,\dots,\hat{y}_T)$ as in the \textsc{Reinforce} algorithm. 
The reward is $r(\tilde{\vec{y}},\vec{y})=r(\hat{\vec{y}},\vec{y})$ once it has the entire trajectory $\hat{\vec{y}}$.

\paragraph{The \textsc{Dagger} Algorithm.}
When $0<\alpha<1$, $\beta=1$, and $\texttt{Decode}(\pi(y\mid\vec{h}_t))$ is defined as as:
\begin{equation}
  \label{eq:greedy}
  \texttt{Decode}(\pi_{\theta}(y\mid \vec{h}_{t-1}))=\argmax_{y'} \pi_{\theta}(y'\mid \vec{h}_{t-1}).
\end{equation}
Depending the value of $\alpha$, $\tilde{y}_{t-1}$ will choose between the ground truth $y_{t-1}$ and decoded value $\hat{y}_{t-1}$ with the function defined in \autoref{eq:greedy}.
On the other hand, $\tilde{y}_t$ will always choose the ground truth $y_t$ as $\beta=1$.
Since $\tilde{\vec{y}}=\vec{y}$, we have $r(\tilde{\vec{y}},\vec{y})=1$ and the reward can be ignored from \autoref{eq:obj}.
In imitation learning, ground truth sequence $\vec{y}$ is called expert actions.
The \textsc{Dagger} algorithm~\citep{ross2011reduction} is also called scheduled sampling~\citep{bengio2015scheduled} in recent deep learning literature.
To be accurate, in the \textsc{Dagger} and the scheduled sampling, the $\alpha$ is dynamically changed during training.
Typically, it starts from 1 and gradually decays to a certain value along with iterations.
As shown in our experiments, the selection of decay scheme has a big impact on model performance.

\paragraph{The \textsc{MLE} Algorithm.}
Besides, there is a trivial case when $\alpha=1, \beta=1$. In this case, $\tilde{y}_{t-1}$ and $\tilde{y}_t$ are equal to $y_{t-1}$ and $y_t$ respectively, and $r(\tilde{\vec{y}},\vec{y})=1$. Optimizing the objective function in \autoref{eq:obj} is reduced to the maximum likelihood estimation (MLE).

\subsection{Other Variant Algorithms}
Inspired by the previous three special cases, we offer other algorithm variants with different combinations of $(\alpha,\beta)$, while the decoding function $\texttt{Decode}(\pi(y\mid\vec{h}_t))$ in the same as \autoref{eq:sampling} in all following variants. 

\begin{itemize}
\item \textbf{\textsc{Reinforce-GTI}} (\textsc{Reinforce} with Ground Truth Input): $\alpha= 1$, $\beta= 0$.
  Unlike the \textsc{Reinforce} algorithm, \textsc{Reinforce-GTI} restricts the input to the decoder can only be ground truth words, which means $\tilde{y}_{t-1} = y_{t-1}$. 
  This is a popular implementation in the deep reinforcement learning for Seq2Seq models~\citep{keneshloo2018deep}.
\item \textbf{\textsc{Reinforce-SO}} (\textsc{Reinforce} with Sampled Output): $\alpha= 1$, $0<\beta<1$.
  In terms of choosing the value of $\tilde{y}_t$ as output from the decoder, \textsc{Reinforce-SO} allows $\tilde{y}_t$ to select the ground truth $y_t$ with probability $\beta$.
\item \textbf{\textsc{Reinforce-SIO}} (\textsc{Reinforce} with Sampled Input and Output): $0<\alpha<1$, $0<\beta<1$.
  Instead of always taking the ground truth $y_{t-1}$ as input, \textsc{Reinforce-SIO} further relaxes the constraint in \textsc{Reinforce-SO} and allows $\tilde{y}_{t-1}$ to be the decoded value $\hat{y}_{t-1}$ with probability $\alpha$. 
\end{itemize}

Unless specified explicitly, an additional requirement when $0<\alpha,\beta<1$ is that its value decays to a certain value during training, which by default is 0.

\section{Experiments}
\label{sec:exp}

\paragraph{Dataset and Evaluation Metrics.}
We evaluate our models on the Quora Question Pair Dataset \footnote{\url{https://www.kaggle.com/c/quora-question-pairs}}, and the Twitter URL Paraphrasing Dataset~\citep{lan-etal-2017-continuously} \footnote{\url{https://languagenet.github.io}}. 
Both datasets contain positive and negative examples of paraphrases, and we only keep the positive examples for our experiments as in prior work of paraphrase generation~\citep{li-etal-2018-paraphrase, patro2018learning}. 
For the Quora dataset, we follows the configuration of~\citep{li-etal-2018-paraphrase} and split the data into 100K training pairs, 30K testing pairs and 3K validation pairs.
For the Twitter dataset, since our model cannot deal with the negative examples as \citep{li-etal-2018-paraphrase} do,  we just obtain the 1-year 2,869,657 candidate pairs from \url{https://languagenet.github.io}, and filter out all negative examples. Finally, we divided the remaining dataset into 110K training pairs, 3K testing pairs and 1K validation pairs.

We use the following evaluation metrics to compare our models with other state-of-art neural networks: \textsc{Rouge-1} and \textsc{Rouge-2}~\citep{lin2004rouge}, \textsc{BLEU} with up to bi-grams ~\citep{papineni2002bleu}.
For the convenience of comparison, we also calculate the average of the scores. 

\paragraph{Competitive Systems.}
We compare our results with four competitive systems on paraphrase generation: the sequence-to-sequence model~\citep[Seq2seq]{bahdanau2014neural}, the Reinforced by Matching framework~\citep[RbM]{li-etal-2018-paraphrase}, the Residual LSTM~\citep[Res-LSTM]{prakash2016neural}, and the Discriminator LSTM model~\citep[Dis-LSTM]{patro2018learning}.
Among these competitive systems, the RbM \citep{li-etal-2018-paraphrase} is more closely related to our work, since we both use the pointer-generator as the base model and apply some reinforcement learning algorithms for policy learning.

\paragraph{Experimental Setup.}
We first pre-train the pointer-generator model with MLE, then fine-tune the models with various algorithms proposed in \autoref{sec:methods}.
Pre-training is critical to make the \textsc{Reinforce} algorithm and some variants to work.
More implementation details are provided in \autoref{sec:details}.

\paragraph{Result Analysis.}

\begin{table*}[t!]
  \centering
  \begin{tabular}{cc|cc|cccc}
    \toprule
    \multicolumn{2}{c|}{}& \multicolumn{2}{c|}{\textsc{Schedule Rate}} & \multicolumn{4}{c}{\textsc{Evaluation Metrics}} \\
    \cmidrule(lr){3-4} \cmidrule(lr){5-8}
    & Models & $\alpha$ & $\beta$ & ROUGE-1 & ROUGE-2 & BLEU & Avg-Score \\
    \midrule 
    1 & Seq2Seq & - & - & 58.77 & 31.47 & 36.55 & 42.26\\
    2 & Res-LSTM & - & - & 59.21 & 32.43 & 37.38 & 43.00\\
    3 & RbM  & - & - & 64.39 & 38.11 & 43.54 & 48.68 \\
    4 & Dis-LSTM  & - & - & - & 44.90 & 45.70 & 45.30\\
    \midrule
    5 & \textsc{Pre-trained MLE}  & $\alpha=1$ & $\beta=1$ & 66.72 & 47.70 & 54.01 & 56.14 \\
    6 & \textsc{Reinforce} & $\alpha=0$ & $\beta=0$ & 67.00 & 47.91 & 54.06 & 56.32 \\
    7 & \textsc{Reinforce-GTI} & $\alpha=1$ & $\beta=0$ & 67.03 & 48.10 & 54.23 & 56.45 \\
    8 & \textsc{Reinforce-SO} & $\alpha=1$ & $\beta \to 0$ & 66.88 & 47.95 & 54.16 & 56.33 \\
    9 & \textsc{Reinforce-SIO} & $\alpha \to 0$ & $\beta \to 0$ & 67.62 & 48.99 & 55.19 & 57.26 \\
    10 & \textsc{Dagger} & $\alpha \to 0$ & $\beta=1$ & 67.64 & 48.96 & 55.06 & 57.22 \\
    11 & \textsc{Dagger*} & $\alpha=0.5$ & $\beta=1$ & \textbf{68.34} & \textbf{49.99} & \textbf{55.75} & \textbf{58.02} \\
    \bottomrule
  \end{tabular}
  \caption{\label{tab:exp}Performance on Quora dataset. The results of competitive systems are reprinted from prior work: line 1 -- 3 are obtained from \citep{li-etal-2018-paraphrase}, line 4 is obtained from \citep{patro2018learning}. The average score in the last column is for the convenience of comparison. }
\end{table*}

 \begin{table*}[t!]
  \centering
  \begin{tabular}{cc|cc|cccc}
    \toprule
    \multicolumn{2}{c|}{}& \multicolumn{2}{c|}{\textsc{Schedule Rate}} & \multicolumn{4}{c}{\textsc{Evaluation Metrics}} \\
    & Models & $\alpha$ & $\beta$ & ROUGE-1 & ROUGE-2 & BLEU & Avg-Score \\
   \midrule 
    1 & \textsc{Pre-trained MLE} & $\alpha=1$ & $\beta=1$  & 58.49 & 43.84 & 38.45 & 46.92 \\
    2 & \textsc{Reinforce} & $\alpha=0$ & $\beta=0$ & 58.67 & 44.06 & 38.46 & 47.06 \\
    3 & \textsc{Reinforce-GTI} & $\alpha=1$ & $\beta=0$ & 58.58 & 43.89 & 38.42 & 46.96 \\
    4 & \textsc{Reinforce-SO} & $\alpha=1$ & $\beta \to 0$ & 58.58 & 43.89 & 38.41 & 46.96 \\
    5 & \textsc{Reinforce-SIO} & $\alpha \to 0$ & $\beta \to 0$ & 58.82 & 44.10 & 38.85 & 47.25\\
    6 & \textsc{Dagger} & $\alpha \to 0$ & $\beta=1$ & 58.84 & 44.24 & 38.95 & 47.34 \\
    7 & \textsc{Dagger*} & $\alpha=0.2$ & $\beta=1$ & \textbf{58.95} & \textbf{44.34} & \textbf{39.04} & \textbf{47.44} \\
    \bottomrule
  \end{tabular}
  \caption{\label{tab:exp_t2}Performance on Twitter dataset. Since the dataset we obtained is different from  \citep{li-etal-2018-paraphrase}, we do not directly compare the results with the prior works.}
\end{table*}

\autoref{tab:exp} shows the model performances on the Quora test set, and \autoref{tab:exp_t2} shows the model performances on the Twitter test set.
For the Quora dataset, all our models outperform the competitive systems with a large margin.
We suspect the reason is because we ran the development set during training on-the-fly, which is not the experimental setup used in \citep{li-etal-2018-paraphrase}.

For both datasets, we find that \textsc{Dagger} with a fixed $(\alpha,\beta)$ gives the best performance among all the algorithm variants.
The difference between \textsc{Dagger} and \textsc{Dagger*} is that, in \textsc{Dagger}, we use the decay function on $\alpha$ at each iteration, $\alpha\leftarrow k\cdot\alpha$ with $k=0.9999$.
In our experiments, we also try different decaying rates, and present the best results we obtained (more details are provided in \autoref{sec:decay}). 
The selection of $\alpha$ depends on the specific task: for the Quora dataset, we find $\alpha = 0.5$ gives us the optimal policy; for the Twitter dataset, we find $\alpha = 0.2$ gives us the optimal policy. 

As shown in line 6 -- 11 from \autoref{tab:exp}, the additional training with whichever variant algorithms can certainly enhance the generation performance over the pre-trained model (line 5).
This observation is consistent with many previous works of using RL/IL in NLP.
However, we also notice that the improvement of using the \textsc{Reinforce} algorithm (line 6) is very small, only 0.18 on the average score.

As shown in line 2 -- 7 from \autoref{tab:exp_t2}, the additional training with variant algorithms also shows improved performance over the pre-trained model (line 1).
However, for the pointer-generator model, it is more difficult to do paraphrase generation on the Twitter dataset. 
Since in the Twitter dataset, one source sentence shares several different paraphrases, while in the Quora dataset, one source sentence only corresponds to one paraphrase.
This explains why the average improvement in the Twitter dataset is not as significant as in the Quora dataset.
Besides, from \autoref{tab:exp_t2}, we also find that IL (line 6 -- 7) outperforms RL (line 2 -- 3), which is consist with the experimental results in \autoref{tab:exp}.

Overall, in this particular setting of paraphrase generation, we found that \textsc{Dagger} is much easier to use than the \textsc{Reinforce} algorithm, as it always takes ground truth (expert actions) as its outputs.
Although, picking a good decay function $\alpha$ can be really tricky. 
On the other hand, the \textsc{Reinforce} algorithm (together with its variants) could only outperform the pre-trained baseline with a small margin.

\section{Related Work}
\label{sec:related}

Paraphrase generation has the potential of being used in many other NLP research topics, such as machine translation~\citep{madnani2007using} and question answering~\citep{buck2017ask,dong2017learning}.
Early work mainly focuses on extracting paraphrases from parallel monolingual texts \citep{barzilay2001extracting,ibrahim2003extracting,pang2003syntax}.
Later, \citet{quirk2004monolingual} propose to use statistical machine translation for generating paraphrases directly.
Despite the particular MT system used in their work, the idea is very similar to the recent work of using encoder-decoder frameworks for paraphrase generation~\citep{prakash2016neural,mallinson2017paraphrasing}.
In addition, \citet{prakash2016neural} extend the encoder-decoder framework with a stacked residual LSTM for paraphrase generation.
\citet{li-etal-2018-paraphrase} propose to use the pointer-generator model~\citep{see2017get} and train it with an actor-critic RL algorithm.
In this work, we also adopt the pointer-generator framework as the base model, but the learning algorithms are developed by uncovering the connection between RL and IL.

Besides paraphrase generation, many other NLP problems have used some RL or IL algorithms for improving performance.
For example, structured prediction has more than a decade history of using imitation learning~\citep{daume2009search,chang2015learning,vlachos2013investigation,liu2018learning}.
In addition, scheduled sampling (as another form of \textsc{Dagger}) has been used in sequence prediction ever since it was proposed in \citep{bengio2015scheduled}.
Similar to IL, reinforcement learning, particularly with neural network models, has been widely used in many different domains, such as coreference resolution~\citep{yin2018deep}, document summarization~\citep{chen2018fast}, and machine translation~\citep{wu2018study}.

\section{Conclusion}
\label{sec:con}

In this paper, we performed an empirical study on some reinforcement learning and imitation learning algorithms for paraphrase generation. 
We proposed a unified framework to include the \textsc{Dagger} and the \textsc{Reinforce} algorithms as special cases and further presented some variant learning algorithms.
The experiments demonstrated the benefits and limitations of these algorithms and provided the state-of-the-art results on the Quora dataset.

\section*{Acknowledgments}
The authors thank three anonymous reviewers for their useful comments and the UVa NLP group for helpful discussion. 
This research was supported in part by a gift from Tencent AI Lab Rhino-Bird Gift Fund.

\bibliography{emnlp-ijcnlp-2019}
\bibliographystyle{acl_natbib}

\clearpage
\appendix

\begin{table*}
  \centering
  \begin{tabular}{c|ccc|cccc}
    \toprule
    \multicolumn{1}{c|}{}& \multicolumn{3}{c|}{\textsc{Schedule Rate}} & \multicolumn{4}{c}{\textsc{Evaluation Metrics}} \\
     & $\alpha$ & $\beta$ & $k_{\alpha}$ & ROUGE-1 & ROUGE-2 & BLEU & Avg-Score \\
    \midrule 
    1 & $\alpha=0.5$ & $\beta=1$ & $k_{\alpha}=1$ & \textbf{68.34} & \textbf{49.99} & \textbf{55.75} & \textbf{58.02} \\
    2 & $\alpha \to 0$ & $\beta=1$ & $k_{\alpha}=0.9999$ & 67.64 & 48.96 & 55.06 & 57.22 \\
    4 & $\alpha \to 0$ & $\beta=1$ & $k_{\alpha}=0.99999$ & 67.92 & 49.45 & 55.44 & 57.60 \\
    4 & $\alpha \to 0$ & $\beta=1$ & $k_{\alpha}=0.999997$ & 67.73 & 49.19 & 55.65 & 57.52 \\
    \bottomrule
  \end{tabular}
  \caption{\label{tab:exp_q}Experiment results for \textsc{Dagger} under different schedule rate settings on Quora dataset}
\end{table*}

\begin{table*}
  \centering
  \begin{tabular}{c|ccc|cccc}
    \toprule
    \multicolumn{1}{c|}{}& \multicolumn{3}{c|}{\textsc{Schedule Rate}} & \multicolumn{4}{c}{\textsc{Evaluation Metrics}} \\
     & $\alpha$ & $\beta$ & $k_{\alpha}$ & ROUGE-1 & ROUGE-2 & BLEU & Avg-Score \\
    \midrule 
    1 & $\alpha=0.2$ & $\beta=1$ & $k_{\alpha}=1$ & \textbf{58.95} & \textbf{44.34} & \textbf{39.04} & \textbf{47.44} \\
    2 & $\alpha \to 0$ & $\beta=1$ & $k_{\alpha}=0.9999$ & 58.84 & 44.24 & 38.95 & 47.34 \\
    4 & $\alpha \to 0$ & $\beta=1$ & $k_{\alpha}=0.99999$ & 58.81 & 44.08 & 38.85 & 47.24 \\
    4 & $\alpha \to 0$ & $\beta=1$ & $k_{\alpha}=0.999997$ & 58.79 & 44.22 & 38.88 & 47.29 \\
    \bottomrule
  \end{tabular}
  \caption{\label{tab:exp_rs_t}Experiment results for \textsc{Dagger} under different schedule rate settings on Twitter dataset}
\end{table*}

\section{Implementation Details}
\label{sec:details}

For all experiments, our model has 256-dimensional hidden states and 128-dimensional word embeddings. Since the pointer-generator model has the ability to deal with the OOV words, we choose a small vocabulary size of 5k, and we train the word embedding from scratch. We also truncate both the input and output sentences to 20 tokens.

For the training part, we first pre-train a pointer generator model using MLE, and then fine-tune this model with the \textsc{Reinforce}, \textsc{Dagger} and other variant learning algorithms respectively. 
In the \textbf{pre-training} phase, we use the Adagrad optimizer with learning rate 0.15 and an initial accumulator value of 0.1; use gradient clipping with a maximum gradient norm of 2; and do auto-evaluation on the validation set every 1000 iterations, in order to save the best model with the lowest validation loss. 
In the \textbf{fine-tuning} phase, we use the Adam optimizer with learning rate $10^{-5}$; use gradient clipping with the same setting in pre-training; and do auto-evaluation on the validation set every 10 iterations. 

When applying the \textsc{Reinforce} and its variant algorithms, we compute the reward as follows:
\begin{equation}
\begin{split}
    r(\tilde{\vec{y}}_n,\vec{y}) &=  \text{ROUGE-2}(\tilde{\vec{y}}_n,\vec{y}) \\
    & - \frac{1}{N}\sum_{n=1}^N \text{ROUGE-2}(\tilde{\vec{y}}_n,\vec{y})
\end{split}
\end{equation}
where $N$ is the total number of sentences generated by random sampling, in all the experiments, we set $N=4$.

At test time, we use beam search with beam size 8 to generate the paraphrase sentence.

\begin{figure}[h]
  \centering
  \includegraphics[width=0.4\textwidth]{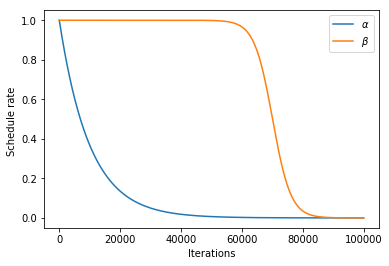}
  \caption{The schedule sampling rate for $\alpha$ and $\beta$}
  \label{fig:schedule_rate}
\end{figure}

According to \citet{bengio2015scheduled}, we define the schedule rate $\alpha^{(i)}= k^i$ (where $0<k<1$, $i$ is the $i$th training iteration), and $\beta^{(i)} = k/(k+exp(i/k))$ (where $k>1$, $i$ is the $i$th training iteration). 
In the experiments shown in Tabel \ref{tab:exp}, for the schedule rate $\alpha$, we set $k=0.9999$; for the schedule rate $\beta$, we set $k=3000$, and the resulting schedule rate curve is shown in Figure \ref{fig:schedule_rate}.

\section{Additional Results}
\label{sec:decay}
\begin{figure}[h]
  \centering
  \includegraphics[width=0.4\textwidth]{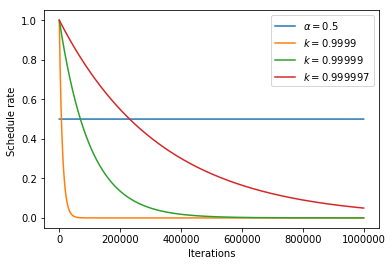}
  \caption{The schedule rate $\alpha$ in \textsc{Dagger}}
  \label{fig:quora_schedule_rate}
\end{figure}

We try different schedule rate settings in \textsc{Dagger} as shown in Figure \ref{fig:quora_schedule_rate}, and compare their model performances on both datasets. The corresponding experimental results are shown in Table \ref{tab:exp_q} and Table \ref{tab:exp_rs_t} respectively. 

We find that if $\alpha$ decreases faster ($k=0.9999$) than the model convergence speed, the model will stop improving before it learns the optimal policy; if $\alpha$ decreases slower ($k=0.999997$) than the model convergence speed, the model will get stuck in the sub-optimal policy. 

For the Quora dataset, we find our model learns the optimal policy when it gets half chance to take the ground truth word $y_{t-1}$ as $\tilde{y}_{t-1}$ (i.e. the schedule rate setting is $\alpha=0.5$ and $\beta=1$).

For the twitter dataset, we find our model learns the optimal policy when it has higher probability to take the decoded word $\hat{y}_{t-1}$ as $\tilde{y}_{t-1}$ (i.e. the schedule rate setting is $\alpha=0.2$ and $\beta=1$).

\end{document}